\documentclass[letterpaper]{article}
\usepackage{graphicx}
\usepackage{caption}
\usepackage{subcaption}
\usepackage{xcolor}
\usepackage{multirow}
\usepackage{amsfonts}
\usepackage{url}

\usepackage[square,numbers]{natbib}
\begin{document}

\newcommand{\theHalgorithm}{\arabic{algorithm}}
%%
%% The "title" command has an optional parameter,
%% allowing the author to define a "short title" to be used in page headers.
\title{DCT: Dual Channel Training of Action Embeddings for Reinforcement Learning with Large Discrete Action Spaces}

\author{
    Pranavi~Pathakota$^*$, Hardik~Meisheri$^*$, Harshad~Khadilkar
}

%%
%% The "author" command and its associated commands are used to define
%% the authors and their affiliations.
%% Of note is the shared affiliation of the first two authors, and the
%% "authornote" and "authornotemark" commands
%% used to denote shared contribution to the research.

%\author{}

% First names are abbreviated in the running head.
% If there are more than two authors, 'et al.' is used.
%

%

%%
%% By default, the full list of authors will be used in the page
%% headers. Often, this list is too long, and will overlap
%% other information printed in the page headers. This command allows
%% the author to define a more concise list
%% of authors' names for this purpose.
%\renewcommand{\shortauthors}{}

%%
%% The abstract is a short summary of the work to be presented in the
%% article.

\maketitle

\def\thefootnote{*}\footnotetext{These authors contributed equally to this work}\def\thefootnote{\arabic{footnote}}

\begin{abstract}

The ability to learn robust policies while generalizing over large discrete action spaces is an open challenge for intelligent systems, especially in noisy environments that face the curse of dimensionality. In this paper, we present a novel framework to efficiently learn action embeddings that simultaneously allow us to reconstruct the original action as well as to predict the expected future state. We describe an encoder-decoder architecture for action embeddings with a dual channel loss that balances between action reconstruction and state prediction accuracy. We use the trained decoder in conjunction with a standard reinforcement learning algorithm that produces actions in the embedding space. Our architecture is able to outperform two competitive baselines in two diverse environments: a 2D maze environment with more than 4000 discrete noisy actions, and a product recommendation task that uses real-world e-commerce transaction data. Empirical results show that the model results in cleaner action embeddings, and the improved representations help learn better policies with earlier convergence.
\end{abstract}

\section{Introduction}

Reinforcement learning (RL) has had significant recent success in applications such as games and robotics \cite{mnih2015human,robotics_RL}. However, real-world problems that involve a large number of discrete action choices are still very challenging for traditional RL algorithms. Examples include scenarios such as recommendation systems \cite{recomm_rl}, supply chains~\cite{meisheri2022scalable}, complex high fidelity games \cite{alphastarblog,berner2019dota}, resource management at scale in data centers \cite{mao2016resource}, investment management \cite{jiang2017deep}, where large action spaces are handled indirectly using pre- or post-processing heuristics. The key challenge is with exploring large action spaces sufficiently well to arrive at optimal policies. Furthermore, hand-crafted heuristics for mapping RL outputs to actions become intractable as the number of actions increases. 

A common reason for explosion of action space is the presence of multiple actuators in a dynamical system (such as a mobile robot \cite{kidzinski2018learning,LearnToRun}), where the number of unique combinations of actions grows exponentially with the number of actuators. A different application but with similar challenges is Recommender Systems, where there are large number of potential actions (product suggestions). Unlike scenarios with fixed rules (such as games) or dynamical equations (robotics), no simple yet effective rules are available for generating recommendations. Existing heuristics that use product features for matching with other similar products typically fail to take into account actual user behaviour and characteristics. However, reinforcement learning algorithms have shown promising results in this area \cite{tang2019reinforcement}, \cite{afsar2022reinforcement}. Unfortunately, the handling of large discrete action spaces (whether in the form of robotic actuators or as product recommendations) is still an open problem.

Recently, the success of state embeddings for complex state spaces has inspired studies on the use of action embeddings along similar lines \cite{DulacArnold2015DeepRL}. The key idea is to learn the RL policy not over raw actions, but over action \textit{representations} in a low dimensional embedding space. If actions with similar effects are grouped close together in the embedding space, the efficiency of exploration is greatly improved. It stands to reason that the better the action representations, the better the chance of reaching good policies.

\begin{figure*}[th]
	\centering
	\begin{subfigure}[b]{0.3\textwidth}
		\centering
		\includegraphics[width=\textwidth]{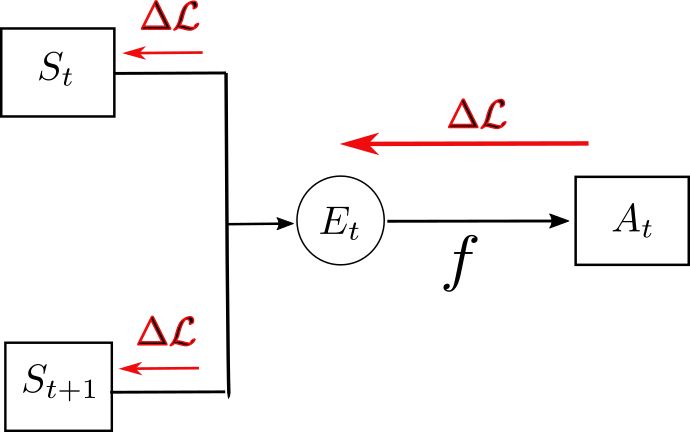}
		\caption{Chandak et. al. \cite{chandak2019learning}}
		\label{fig:yash_arch}
	\end{subfigure}
	\hfill
	\begin{subfigure}[b]{0.3\textwidth}
		\centering
		\includegraphics[width=\textwidth]{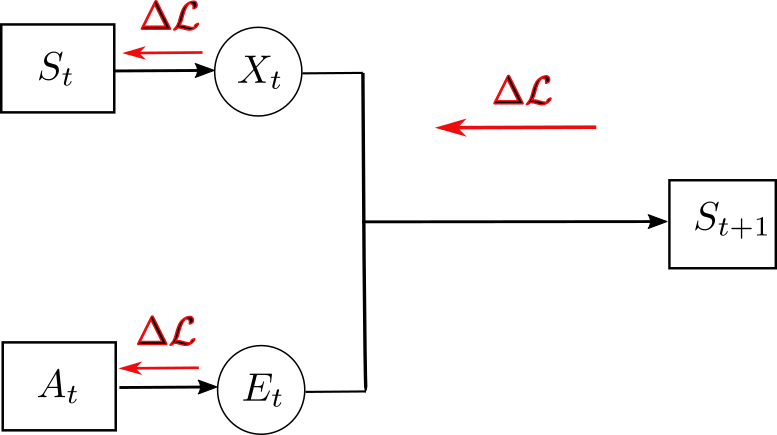}
		\caption{Pritz et. al. \cite{joint_training}}
		\label{fig:paul_arch}
	\end{subfigure}
	\hfill
	\begin{subfigure}[b]{0.3\textwidth}
		\centering
		\includegraphics[width=\textwidth]{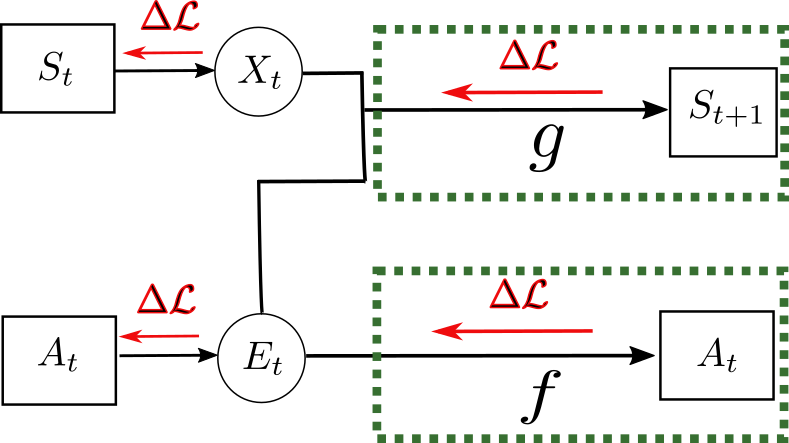}
		\caption{Proposed Architecture}
		\label{fig:our_arch}
	\end{subfigure}
	\caption{Comparison with prior work. The proposed Dual Channel Training (DCT) architecture improves learnt embeddings $E_t$ with a significant effect on the rate and quality of policy learning, as shown in this paper.}
	\label{fig:comp_arch}
\end{figure*}

In this paper, we present an architecture to efficiently learn action embeddings in low dimensional space. We force the embeddings to be rich by imposing the dual task of learning the effect of actions as well as predicting future states. We show experimentally that this helps the RL agents learn better policies in scenarios with large action spaces. We build upon work of Chandak et al. \cite{chandak2019learning} and Pritz et al. \cite{joint_training} and provide a generalized framework for learning embeddings which is not only efficient in encoding transition dynamics between states but also helps in decoding those actions into actual spaces (Fig. \ref{fig:comp_arch}). A detailed review is given in the next section. We test our model on an environment which is of stochastic nature and test our method up to ${2^{12}}$ discrete actions. We experimentally show that our method outperforms \cite{chandak2019learning,joint_training} as well as a standard DQN baseline in almost all cases. In a second experiment, we compare our algorithm with the same two baselines on a recommendation dataset obtained from real-world e-commerce transactions. We show that the proposed algorithm is able to outperform the baselines in terms of the cosine similarity between the embedding of the recommendation and the embedding of the actual product purchased. 

The main contributions of our work are as follows:

\begin{itemize}
    \item We propose a new architecture for an action encoder-decoder model which can be trained simultaneously (both the embedding to low-dimensional space and the recovery of actions in the original space). It results in a better representation of action embeddings by jointly training encoder and decoder for action reconstruction and next state prediction.

    \item We present extensive experimentation over a \textit{noisy} maze environment with up to $2^{12}$ unique actuator actions to validate our model and compare it with previous work and a traditional off policy RL algorithm (DQN).

    \item We also test our proposed algorithm on a practical application of recommender systems. We show that we are able to outperform the baselines on a real-world fashion e-commerce data set.
\end{itemize}

The remainder of this paper is structured as follows: In Section 2, give an overview of related work. The preliminaries and notations are introduced in Section 3. The proposed methodology is presented in Section 4, followed by evaluations in Section 5.

\section{Related Work}

Numerous studies have proposed the use of latent representations of actions in order to solve reinforcement learning (RL) problems. One of the first studies to describe the problem \cite{DulacArnold2015DeepRL} presented a technique for learning a policy over action representations to handle exploding discrete action spaces. They assume that action representations are already known, while the RL policy outputs a `proto-action' in embedding space and uses k-nearest-neighbor search to decode the action representation to actual action. A similar assumption of prior embedding knowledge is present in \cite{tennenholtz2019natural}, followed by policy training using expert demonstrations.

Other studies have focused on the use of action embeddings for task transfer \cite{chen2019learning} and later for generalising to unseen actions \cite{jain2020generalization}. In both cases, the primary objective was to learn embeddings that are well suited to generalise on a distribution of related tasks. They also depended on knowledge of a hierarchical structure in the action space (for example, through the use of a hierarchical variational autoencoder \cite{jain2020generalization}). The autoencoder structure has also been used for generating embeddings for robotic applications \cite{losey2020controlling}, again with the focus on solving a specific task. An alternative perspective is to handle environments where certain actions are redundant \cite{baram2021action}, i.e., have identical effects on the environment. In each of these studies, the embeddings are learnt with the sole objective of improving policy learning, and there are no specific requirements on the quality of the embeddings. As we shall see, these characteristics do seem to matter in our experiments. Additional evidence is provided by results from recent work \cite{hua2022simple} that indicates improved performance when action and state representations are concatenated in multi-task environments. In the present paper, we focus on capturing the semantics and structure of action representations as they affect the environment, independent of the task to be performed.

%{\color{red}
%Recently, in similar lines, emergent action representations \cite{hua2022simple} is proposed, which learns embeddings for multi tasks assuming that these tasks share the same latent action representation space. An action decoder which is shared across different tasks is learnt by integrating time-varying sensory component(state representation) which is used to encode the same modalities across the tasks and time-invariant component(action representation) specific to that task, which resulted in efficient task adaptation. Our proposed approach can also be adapted to multiple tasks provided they share the similar dynamics, but our main focus is to capture the semantics and structure of the action representations focusing for a single task.}

The hierarchical theme is also present in Hybrid RL \cite{fan2019hybrid}, where the action space is split into discrete subsets of parameterised actions. The branching of layers in hierarchical action structure can include discrete or continuous action spaces. However this approach works only if we already know the hierarchical structure of action space and can classify similar actions at different levels in the tree. Another approach for pruning the exploration of action space is action elimination \cite{zahavy2018learn}, which restricts the search to ``optimal'' actions only. However, these action sets have to be externally defined. Apart from instantaneous action representations, one prior study has considered embeddings of temporal sequences of actions \cite{whitney2019dynamics}. They use a VAE for generating both state and action representations. The encoder is trained such that similar action sequences have embeddings close to each other. The decoder maps the action representation from a policy to sequence of high level actions. However, the actions are assumed to be in the continuous domain, with obvious definitions of (dis)similarity. In this work, we focus on large discrete action spaces.

Perhaps the two closest studies to the proposed work are the ones mentioned in the previous section. Chandak et al \cite{chandak2019learning} train action embeddings with the current and next state as the input, with labels provided by the actions that enable the specified transition (Fig. \ref{fig:comp_arch}-(a)). The RL policy outputs actions directly in the embedding space $E_t$, and the embedding to action function $f$ (learnt or distance measure) is used to compute the true action. Pritz et al \cite{joint_training} use a similar concept but with two changes. First, they use the current state and action as the input context, and train the embeddings $E_t$ with respect to the prediction loss for the next state. Second, they also train a latent representation $X_t$ of the state input (Fig. \ref{fig:comp_arch}-(b)). Given the lack of a loss based on the distinction between embeddings of similar actions, there is a good chance of overlap between action embeddings. 

\begin{figure}[t]
     \centering
         \includegraphics[width=0.5\textwidth]{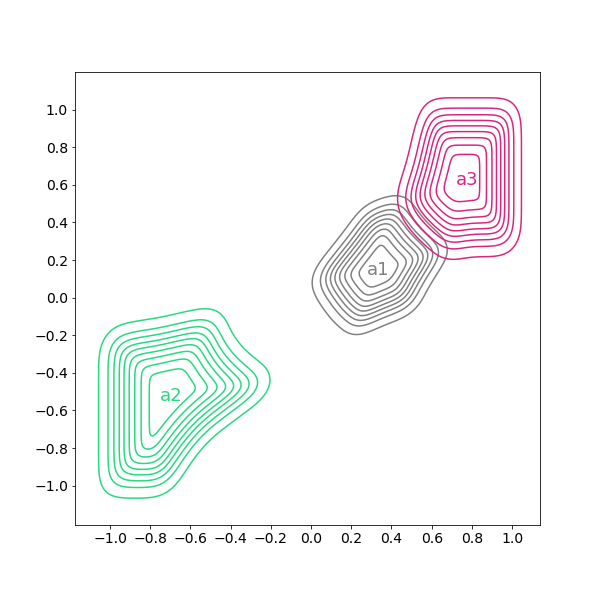}
         \caption{Contour plot of probability of inverse mapping function. We assume (for illustration) that $a_1$ and $a_3$ have similar effects on the environment, while $a_2$ has a very distinct effect. Decoding will be difficult in the region of overlap.}
         \label{fig:cont_plot_prob_actions}
\end{figure}

Fig. \ref{fig:cont_plot_prob_actions} demonstrates this point intuitively. We assume that actions $a_1$ and $a_3$ correspond to similar transitions (perhaps in different parts of the state space) while $a_2$ is very distinct. The Pritz et al \cite{joint_training} architecture will embed $a_2$ in a separate region, by design. However, the regions for $a_1$ and $a_3$ could overlap (there is no intrinsic incentive for distinct embeddings), leading to difficulty during decoding. Unless they have some separation boundary, the decoder $f$ cannot map those continuous values to the original discrete action space. 

In our proposed architecture, the Dual Channel Training (DCT) loss ensures that the action embeddings $E_t$ encode information sufficient for both action reconstruction (decoder $f$ in Fig. \ref{fig:comp_arch}-(c)) as well as next-state prediction (function $g$). This ensures that we keep track of the environment dynamics as well as on distinct action embeddings. As we show in Section \ref{sec:results}, DCT shows a clear improvement in sample efficiency and policy quality due to this change.

\section{Preliminaries}

We model a generic reinforcement learning problem as a Markov Decision Process (MDP) as per \cite{sutton2018reinforcement}, consisting of finite (but large) state and action spaces. Further, we assume in this work that the action space is also discrete. An MDP is defined by a tuple $(\mathcal{S}, \mathcal{A}, \mathcal{T}, \mathcal{R}, \gamma, \rho_{o})$. The state, action, and reward at time $t \in \mathbb{Z}^+$ are denoted by the random variables $s_t\in\mathcal{S}$, $a_t\in\mathcal{A}$, and $r_t\in\mathcal{R}$. $\mathcal{T}$ represents the transition function from one state to another and parameterised by the action, defined by $\mathcal{T}: \mathcal{S} \times \mathcal{A} \rightarrow \mathcal{S}$. The transition function can be deterministic or stochastic. The discount factor $\gamma \in $ [0, 1] applies to future rewards and $\rho_{o}$ represents the distribution of initial states. The goal of the agent is to find an optimal policy that maximizes the expected future return  $\mathbb{E}[\sum_{t=0}^{\infty} \gamma^{t} r_{t}]$. 
\begin{figure}[t]
		\centering
		\includegraphics[width=0.4\textwidth]{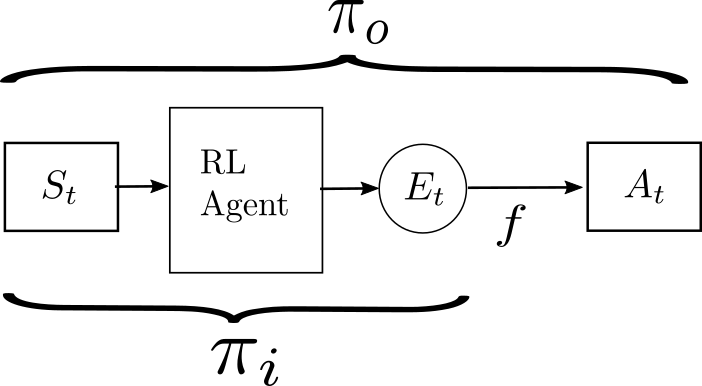}
		\caption{Composite outer policy $\pi_o=f(\pi_i(s))$ composed of an internal policy $\pi_i$ (trained in the RL phase) and a decoder function $f$ (learned during the encoder-decoder training).}
		\label{fig:rl_training}
\end{figure}

%\subsection{Problem Formulation}
Our work is mainly inspired from \cite{chandak2019learning} and \cite{joint_training} where action representations are learnt for projecting high dimensional discrete actions onto a low dimensional continuous domain. The key idea behind generalization in this context is to capture the underlying structure of these actions as well as the environment dynamics which are influenced by the actions. As mentioned earlier, there are parallels to state embeddings which are well studied in literature and have shown benefits in policy optimization. Instead of optimizing over a large discrete space, it is optimized over low dimensional continuous representation space $\varepsilon \subseteq \mathbb{R}^{d}$. 

As shown in Fig. \ref{fig:rl_training}, we aim to learn a composite outer policy $\pi_o$ in two phases. In the first phase, we learn a parameterised embedding-to-action mapping function 
$f:\varepsilon \rightarrow \mathcal{A}$ via encoder-decoder training (see Fig. \ref{fig:comp_arch}). In the second phase, we learn an internal policy $\pi_i:\mathcal{S}\rightarrow\varepsilon$ such that the distribution of $A_t$ is characterised by,

\begin{equation}
	E_t \approx \pi_{i}(.|S_t), \;\;\; A_t = f(E_t), \;\;\; A_t \approx \pi_o(.|S_t)=f(\pi_i(.|s_t))
\end{equation}
Clearly, there are many regimes in which we can optimise $\pi_o$. The simplest option is the two-phased approach described above, with training of $f$ followed by optimisation of $\pi_i$. The composite function $\pi_o$ subsumes cases where state $s_t$ is given directly or via state embeddings \cite{chandak2019learning,joint_training}. We can also consider iterative training of $f$ and $\pi$ among various other options, though we do not follow this up in the present work. Our working hypothesis is that the two-phased approach with $f$ trained using DCT is itself a significant improvement over present baselines.

%And optimizing $\pi_{i}$ leads to optimizing of $\pi_{o}$ with fixed as well learned $f$. This is also true if states are also converted to their respective state embeddings \cite{chandak2019learning,joint_training}. However, we hypothesize that when input is state and action as mentioned in \cite{joint_training} and optimizing only for the state transition model not efficient information encoding into the action embeddings. Specifically, learning embedding-to-action mapping function "f" along with the learning transition function helps in having much better encoding of embeddings.

\section{Methodology} \label{sec:method}
In this section, we propose a model to efficiently learn action embeddings using an encoder-decoder architecture with Dual Channel Training (DCT). We focus on the explanation of action embeddings $E_t$ from Fig. \ref{fig:comp_arch}, but an analogous method\footnote{For state embeddings, we only use the gradient from the next-state prediction loss} can be used to train state embeddings $X_t$. Following this step, we can use any off-the-shelf model-free RL algorithm to train the internal policy $\pi_i$. We use DDPG \cite{silver2014deterministic} for most experiments in this paper.

As depicted in Fig.~\ref{fig:comp_arch}, our model trains embeddings using an encoder-decoder framework. Given state $s_{t}$ and action $a_{t}$, the encoder portion gives embeddings $X_t$ and $E_t$ respectively. The decoder consists of two channels, parameterised by functions $g$ and $f$. Inverse mapping function $f$ maps the dense representation $E_t$ to recover the actual action $a_t$. The transition model $g$ predicts $s_{t+1}$ given the state and action embeddings. Note that $f$ depends only on $E_t$ as input while $g$ takes into account both $X_t$ and $E_t$ to reconstruct the next state. It is worth noting that if the states are linear they can be reconstructed back to its original space ($X_t$ is the identity mapping). If the states are complex (such as images), then reconstruction can be done in embedding or latent space ($X_t$ is a learnt dense representation) as done in \cite{meisheri2021folar}.

The encoder-decoder model is jointly trained using DCT, with loss gradients flowing through both $f$ and $g$. The generic loss function is given by,
\begin{equation}
    %\mathcal{L} = L_1(\hat{s}_{t+1}, s_{t+1}) - \eta \times \frac{1}{N}\sum_{i=1}^{N} \left[ j_i\,\log(f(E_t)) \right],
    \mathcal{L} = \underbrace{L_1(g(X_t,E_t), s_{t+1})}_{\mathrm{prediction}\,\mathrm{loss}} -\, \eta \times \underbrace{\frac{1}{N} \times \log P(a_t|f,E_t)}_{\mathrm{reconstruction}\,\mathrm{loss}} , \label{eq:loss}
\end{equation}
where $L_1$ is a metric to measure the state prediction loss, $N$ is the number of actions, and $P(a_t|f,E_t)$ is the softmax probability of decoding the embedding $E_t$ to the correct action $a_t$, as parameterised by $f$. As mentioned earlier, the prediction loss can be defined directly over the state $s_{t+1}$ (through KL divergence or mean squared error) if the state space is linear, or we can use the embedding $X_{t+1}$ if the state space is complex. The action reconstruction loss is a simple cross-entropy term where the truth vector is one-hot encoded for the input action $a_t$, and the output probability distribution is the softmax distribution produced by the trained decoder $f$. The multiplier $\eta$ is a hyperparameter used for trading off the importance between the two loss terms.

After the encoder-decoder model has been trained, the internal policy $\pi_i$ is learnt over the low dimensional action embedding space $\varepsilon$ using any standard model-free RL algorithm. Encoder function is no longer used, while the decoder $f$ is used to map the output action representations from $\pi_i$ to the original action space as shown in Fig. \ref{fig:rl_training}. 

\section{Results: Naviation in 2D Maze} \label{sec:results}

We present results on a 2-D Maze environment, which an agent with a number of directional actuators is expected to navigate. Apart from performance comparisons, we also investigate the effect of the embeddings and hyperparameters, and provide intuition about the differences in performance.

\subsection{Maze Environment}

\begin{figure}[h]
	\centering
	\includegraphics[width=0.37\textwidth]{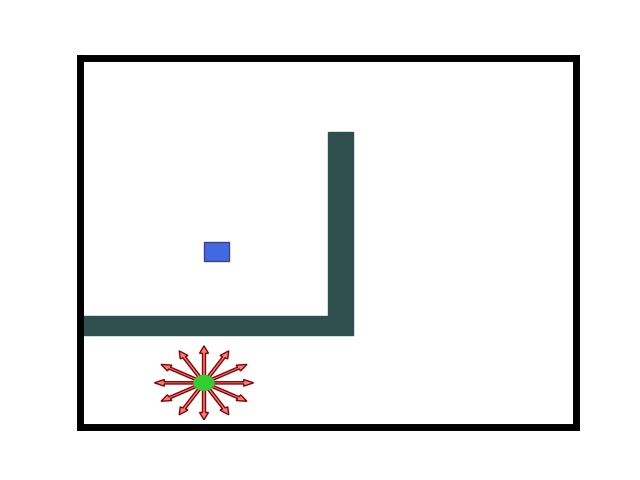}
	\caption{Maze Environment: Green dot at the bottom denotes agent position. Actuator directions are represented by red arrows around the agent. Blue square is the goal state, and obstacles are denoted by grey walls.}
	\label{fig:env_grid_world}

\end{figure}

We trained and tested the proposed architecture on a maze environment designed by Chandak et al \cite{chandak2019learning}, which is also a close baseline of this work. The environment incorporates an agent which has $n$ actuators with equal angular spacing around it. Figure~\ref{fig:env_grid_world} provides an illustration of the environment.  Actuators can either be turned on or off, and the action $a_t$ is the binary string of $n$ digits that describes the setting of each actuator. Clearly, the action space scales exponentially with the number of actuators, with $2^n$ unique actions for $n$ actuators. When the actuator is turned on, it would cause the agent to move in the direction it is pointing towards in the 2D space. The final displacement of the agent would be the sum of displacement vectors of all the selected actuators. 

The state space comprises of Cartesian coordinates $(x_{a,t},y_{a,t})$ of the agent and the coordinates $(x_g,y_g)$ of the goal position. At each time step $t$, the agent selects the binary setting of each of the $n$ actuators. A small penalty of $-0.05$ is given for each intermediate step, and a terminal reward of $+100$ is provided for reaching the goal. In addition, penalty of $-0.01$ is given if agent collides with any obstacle. A timeout of $500$ steps is applied in case the agent does not reach the goal. There are obstacles present in the environment (grey walls in Fig.~\ref{fig:env_grid_world}), which restrict the movement of the agent. This also provides exploration challenges as the agent has to travel around the walls to reach the goal. Additional noise is provided by ignoring the specified action with a probability of $0.1$, and applying a uniform random action instead.

The randomness during the course of learning and the exploding size of the action space make the environment a highly complex stochastic problem. Also, as there are many combinations of selecting actuators that result in a similar final displacement vector output for the agent, this environment acts as a perfect simulator to leverage the structure present in the actions and represent them in a low dimensional space.

\subsection{Baselines and RL training}

For RL training, we have used Deep Deterministic Policy Gradient (DDPG) \cite{silver2014deterministic} with soft updates for the internal policy $\pi_i$ for all three embedding approaches (the proposed approach, PG-RA by Chandak et al \cite{chandak2019learning}, and JSAE~\footnote{Codebase for JSAE was not available hence we have trained using implementation details available in the paper} by Pritz et al \cite{joint_training}). We extend the DDPG algorithm to work in a 2-dimensional action embedding space, instead of single continuous action. Note that the approach should work similarly with any other model-free RL algorithm as well. Since DDPG is an off-policy algorithm, we compare it with DQN~\cite{mnih2015human} as a third baseline. The DQN baseline has no action embeddings, and uses the raw state $s_t$ as input with $2^n$ outputs for the actions.

For the proposed approach as well as the two baselines (except DQN), we first train the embeddings using the respective architectures, and then train $\pi_i$ using DDPG. All the results presented here are averaged across 10 random seeds. Since for maze environment, state is denoted by Cartesian coordinates, we have used mean squared loss for $L1$. Our neural network architecture has 4 layers for generating $E_t$. For action space of size $2^n$, we have a fully connected architecture with layers of size [$2^n$/2, $2^n$/4, $2^n$/8, 2] neurons, with last layer denoting the use of a 2-D embedding space $\varepsilon$. The decoder function $f$ has an exactly inverse architecture for recovering the original action. We have used \texttt{adam} optimizer for training with batch size of $128$. For training $\pi_i$ using DDPG, the actor and critic both use a 4 layer network with [$30, 20, 10$] neurons up to the penultimate layer, followed by output layer of 1 neuron for the critic and 2 neurons for the actor. We have used \texttt{tanh} activation throughout all the networks which helps in restricting the values of $E_t$ in addition to having good gradient properties. Since state space in the maze environment is linear we have not used state embeddings $X_t$ for DCT. 

\subsection{Training Results}

%Figure~\ref{fig:training_res} presents the training results for $2^{10}$ and $2^{11}$ actions (10 and 11 actuators respectively). We can observe that DCT outperforms all the other baseline algorithms, converging earlier and reaching a higher reward. In addition, it is worthwhile to note that DCT has a significantly lower standard deviation across random seeds, especially in larger action spaces providing evidence of the robustness of embeddings. We also conducted experiments from $2^{4}$ to $2^{12}$ actions to analyze the consistency and trend for varying action space sizes. Results are presented in Fig.~\ref{fig:training_res}. As the number of actions increase, DQN is not able to perform well and its performance drastically decreases after $2^7$ actions. We can also observe that JSAE and PG-RA are not able to perform consistently across the different number of actions. We believe that more is exploration needed for the policy to optimize and understand the embedding structure. We can see that even with just $500$ episodes (Fig.~\ref{fig:training_res}), DDPG over DCT embeddings is able to learn consistently across the actions. 

\begin{figure}[h]
     \centering
     \begin{subfigure}[b]{0.48\textwidth}
         \centering
         \includegraphics[width=\textwidth]{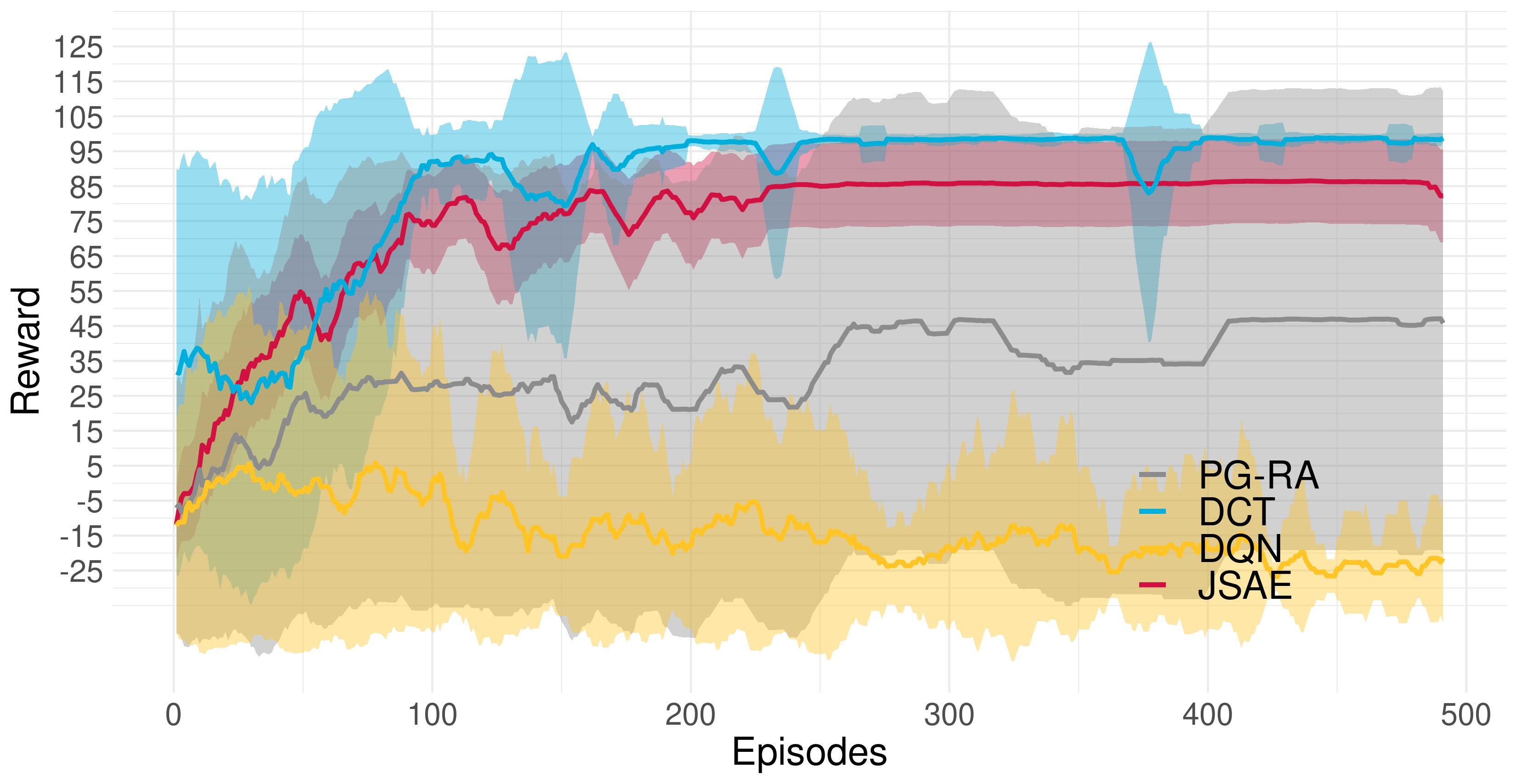}
         \caption{$2^{10}$}
     \end{subfigure}
     \hfill
     \begin{subfigure}[b]{0.48\textwidth}
         \centering
         \includegraphics[width=\textwidth]{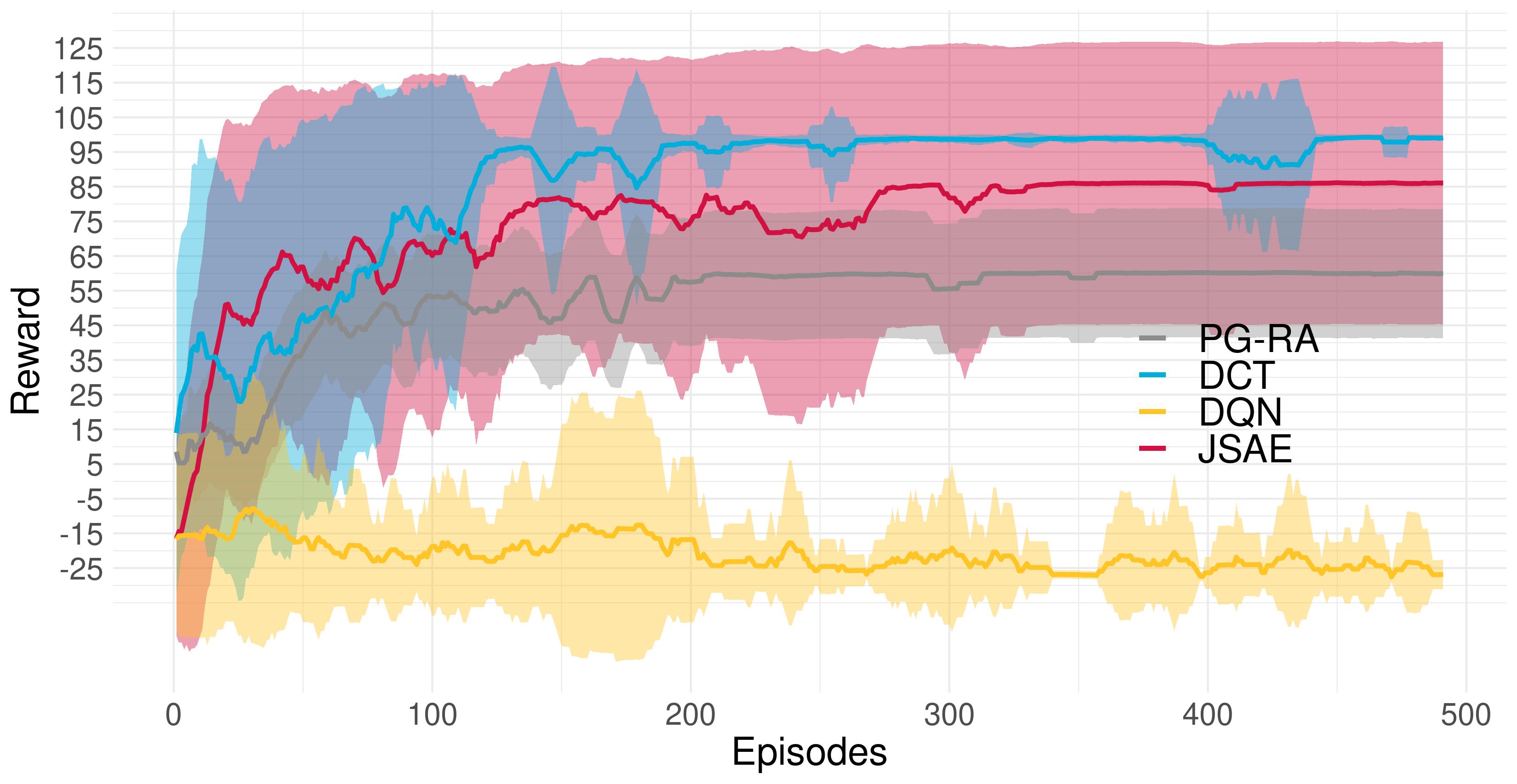}
         \caption{$2^{11}$}
     \end{subfigure}
     \hfill
    \caption{Training results for $2^{10}$ and $2^{11}$ actions: DCT in blue, JSAE in red, PG-RA in grey, and DQN in yellow. Averages over 10 random seeds.}
    \label{fig:training_res}
        
\end{figure}

Figure~\ref{fig:training_res} presents the training results for $2^{10}$ and $2^{11}$ actions (10 and 11 actuators respectively). We can observe that DCT outperforms all the other baseline algorithms, converging earlier and reaching a higher reward. In addition, it is worthwhile to note that DCT has a significantly lower standard deviation across random seeds, especially in larger action spaces providing evidence of the robustness of embeddings.

We also conducted experiments from $2^{4}$ to $2^{12}$ actions to analyze the consistency and trend for varying action space sizes. Results are presented in Table~\ref{tab:training_res} with mean and standard deviations calculated across 10 random seeds. As the number of actions increases, performance of DQN degrades and drastically decreases after $2^6$ actions. We can also observe that PG-RA is not able to perform consistently across the different number of actions. We believe that more exploration is needed for the policy to optimize and understand the embedding structure. We can see that even with just $500$ episodes (Fig.~\ref{fig:training_res}), DDPG over DCT embeddings is able to learn consistently across the actions.

\begin{table*}[h]
\centering
\caption{Training Results}
\label{tab:training_res}
\resizebox{0.9\textwidth}{!}{%
\begin{tabular}{|l|ll|ll|ll|ll|ll|}
\hline
\multirow{2}{*}{}    & \multicolumn{2}{c|}{$2^4$}          & \multicolumn{2}{c|}{$2^6$}          & \multicolumn{2}{c|}{$2^8$}          & \multicolumn{2}{c|}{$2^{10}$}         & \multicolumn{2}{c|}{$2^{12}$}         \\
                     & \multicolumn{1}{c|}{$\mu$} & \multicolumn{1}{c|}{$\sigma$} & \multicolumn{1}{c|}{$\mu$} & \multicolumn{1}{c|}{$\sigma$} & \multicolumn{1}{c|}{$\mu$} & \multicolumn{1}{c|}{$\sigma$} & \multicolumn{1}{c|}{$\mu$} & \multicolumn{1}{c|}{$\sigma$} & \multicolumn{1}{c|}{$\mu$} & \multicolumn{1}{c|}{$\sigma$} \\ \hline
DQN                  & \multicolumn{1}{l|}{38.74}  & 58.72                   & \multicolumn{1}{l|}{97.38} & 4.05                    & \multicolumn{1}{l|}{3.416} & 48.86                   & \multicolumn{1}{l|}{-22.03} & 12.24                   & \multicolumn{1}{l|}{-25.92} & 4.79                    \\
PG-RA                   & \multicolumn{1}{l|}{25.13}  & 65.10                   & \multicolumn{1}{l|}{97.46} & 0.46                    & \multicolumn{1}{l|}{9.70}  & 61.74                   & \multicolumn{1}{l|}{45.91} & 65.31                   & \multicolumn{1}{l|}{73.15}     & 53.31                    \\
JSAE                 & \multicolumn{1}{l|}{97.89} & 1.32                    & \multicolumn{1}{l|}{\textbf{98.76}} & 0.33                   & \multicolumn{1}{l|}{\textbf{98.78}} & 0.78                   & \multicolumn{1}{l|}{82.25} & 39.05                    & \multicolumn{1}{l|}{90.04} & 27.02                   \\
DCT                  & \multicolumn{1}{l|}{85.53}  & 40.56                   & \multicolumn{1}{l|}{72.69} & 54.03                   & \multicolumn{1}{l|}{\textbf{98.13}} & 1.492                   & \multicolumn{1}{l|}{\textbf{98.39}}  & 1.09                    & \multicolumn{1}{l|}{\textbf{99.15}}  & 0.15                    \\
DCT-Euc            & \multicolumn{1}{l|}{\textbf{98.31}}       &  0.91                       & \multicolumn{1}{l|}{85.43}      &   38.45                      & \multicolumn{1}{l|}{\textbf{99.19}}      &  0.19                       & \multicolumn{1}{l|}{\textbf{99.03}}       &   0.29                      & \multicolumn{1}{l|}{60.40}       &   58.19                      \\ \hline
%DCT-SC-Euc & \multicolumn{1}{l|}{97.79}       & 1.08                         & \multicolumn{1}{l|}{97.67}      &     3.96                   & \multicolumn{1}{l|}{98.61}      &    0.99                     & \multicolumn{1}{l|}{86.19}       &   38.69                      & \multicolumn{1}{l|}{98.75}       &     0.51                    \\ \hline
\end{tabular}%
}
\end{table*}

Tab.~\ref{tab:training_res} comprehensively presents the results among the baselines and DCT. DCT-Euc represents the variation of DCT when instead of the learned inverse mapping function $f$, we use Euclidean distance among the embeddings of actions and policy output. PG-RA and JSAE both use Euclidean distance (nearest neighbours) as the inverse mapping function from embedding space to original actions. We can observe that both DCT and DCT-Euc outperform all the other baselines. In addition, as the number of actions increases DCT with the learned $f$ function is able to perform much better. We believe that this is because of the limited size of the embedding space, favouring embeddings with cleaner separation. This is especially noticeable in the performance drop for DCT-Euc with $2^{12}$ actions.
%this is due to complex decision boundaries present in the embedding space as the number of actions increases. 

\begin{figure*}
	\centering
	\begin{subfigure}[b]{0.24\textwidth}
		\centering
		\includegraphics[width=\textwidth]{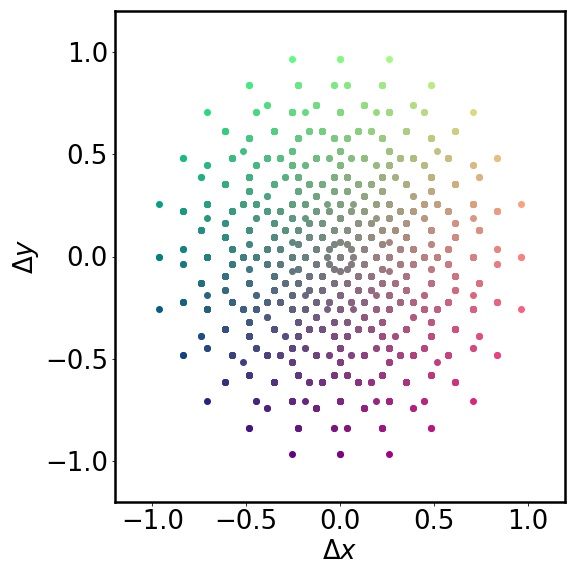}
		\caption{}
	\end{subfigure}
	\hfill
	\begin{subfigure}[b]{0.24\textwidth}
		\centering
		\includegraphics[width=\textwidth]{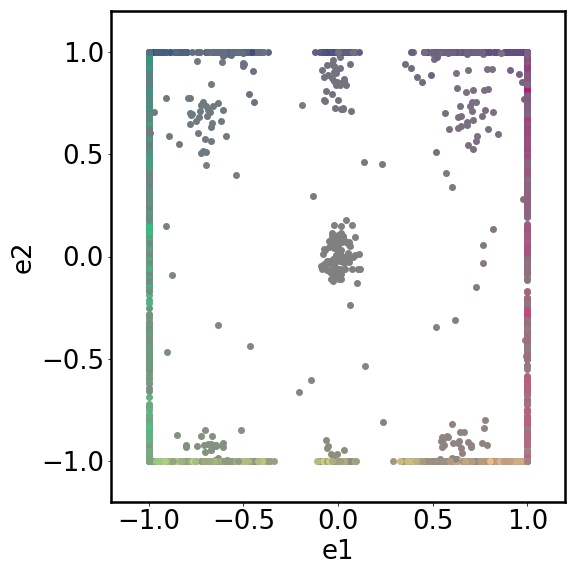}
		\caption{}
	\end{subfigure}
	\hfill
	\begin{subfigure}[b]{0.24\textwidth}
		\centering
		\includegraphics[width=\textwidth]{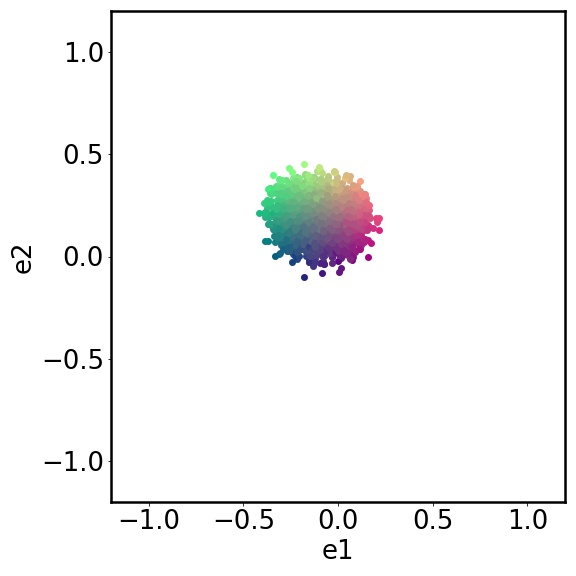}
		\caption{}
	\end{subfigure}
	\hfill
	\begin{subfigure}[b]{0.24\textwidth}
		\centering
		\includegraphics[width=\textwidth]{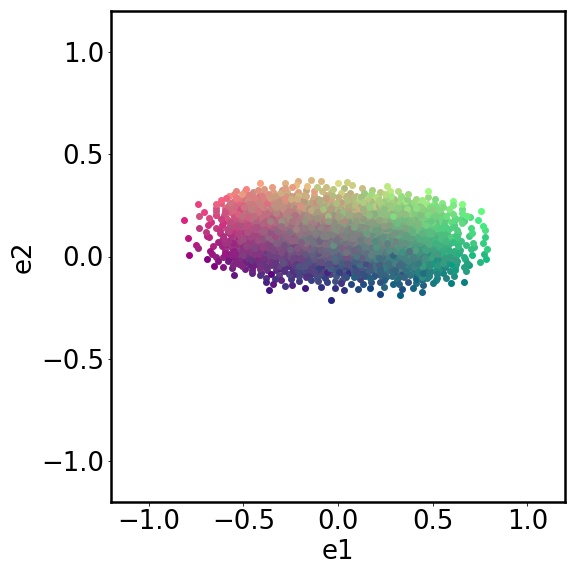}
		\caption{}
	\end{subfigure}
	\caption{Visualization of a) actual displacement on the x-axis and y-axis in the environment, b, c, d) embeddings learned using PG-RA, JSAE and DCT respectively. Each action is colour graded with RGB with R = $\Delta x$, G = $\Delta y$, B = 0.5}
	\label{fig:act_12_viz}
	
\end{figure*}

Figure~\ref{fig:act_12_viz} presents a visualization for $2^{12}$ number of actions. As compared to PG-RA and JSAE, we can observe that DCT embeddings provide more separation between actions and thus help in decoding function $f$. PG-RA embeddings are much more spread across the whole space however they are more concentrated on the boundaries with large empty spaces in the centre. This usually presents a challenge during exploration for policies as a large number of points in the embeddings space would be getting to single or few actions. On the other hand, JSAE embeddings do retain the structure of the true displacement but they are squeezed into much smaller space which also hampers exploration for the RL algorithm. 

\subsection{Effect of entropy coefficient $\eta$}

\begin{figure*}[h]
	\centering
	\begin{subfigure}[b]{0.24\textwidth}
		\centering
		\includegraphics[width=\textwidth]{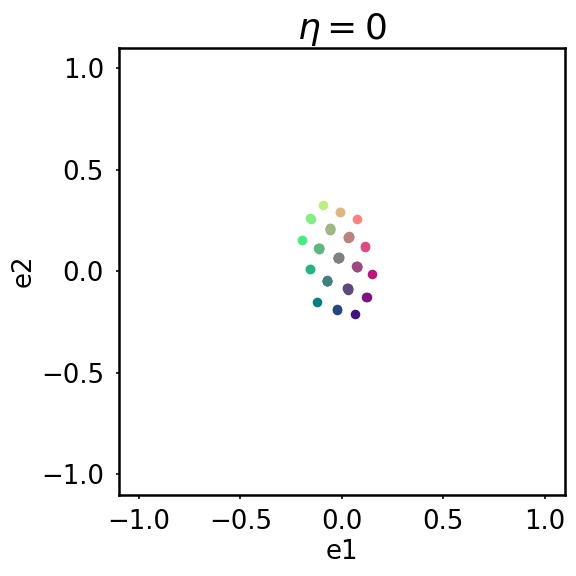}
		%\caption{Cartesian Coordinates}
		\label{fig:$SA_NO_GS_coef_0 6 actions$}
	\end{subfigure}
	\hfill
	\begin{subfigure}[b]{0.24\textwidth}
		\centering
		\includegraphics[width=\textwidth]{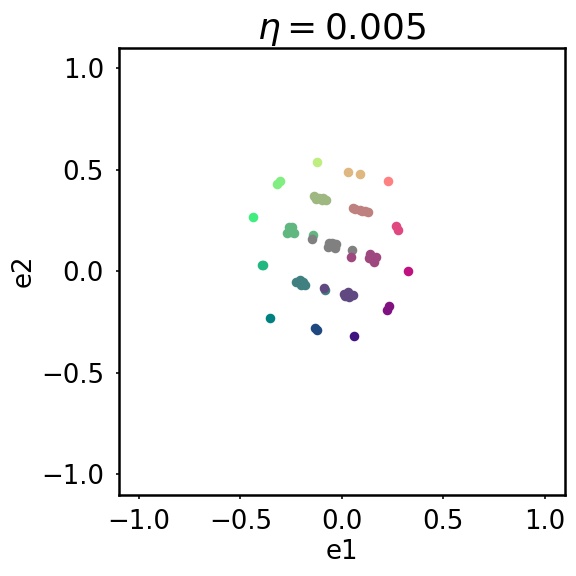}
		%\caption{Cartesian Coordinates}
		\label{fig:$SA_NO_GS_coef_5 6 actions$}
	\end{subfigure}
	\hfill
	\begin{subfigure}[b]{0.24\textwidth}
		\centering
		\includegraphics[width=\textwidth]{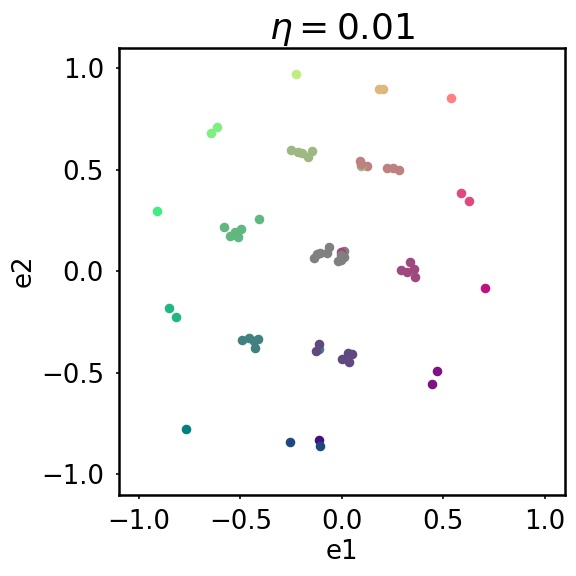}
		%\caption{SA Representations}
		\label{fig:$SA_NO_GS_coef_10 6 actions$}
	\end{subfigure}
	\hfill
	\begin{subfigure}[b]{0.24\textwidth}
		\centering
		\includegraphics[width=\textwidth]{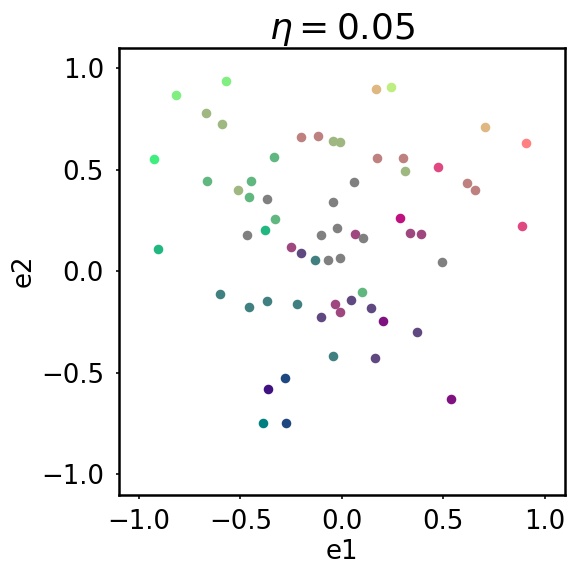}
		%\caption{SNS Representations}
		\label{fig:$SA_NO_GS_coef_50 6 actions$}
	\end{subfigure}
	\hfill
	\caption{Effect of $\eta$ for $ 2^{6} $ actions}
	\label{fig:effect_eta}
	
\end{figure*}

\begin{figure}[h]
	\centering
	\includegraphics[width=0.45\textwidth]{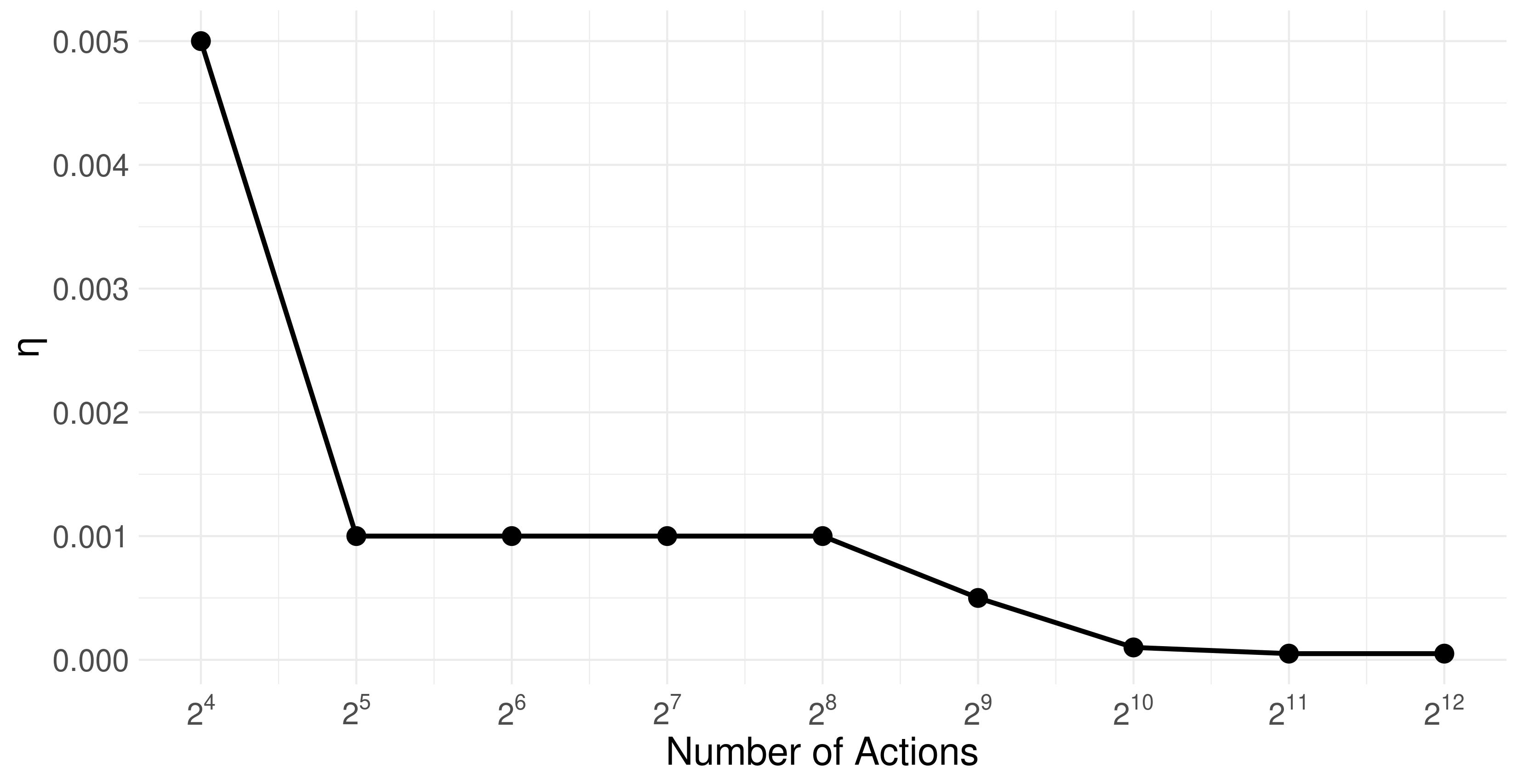}
	\caption{Value of $\eta$ for which we get maximum spread without losing out on structure of embedding}
	\label{fig:eta_agg}
\end{figure}

As mentioned earlier, $\eta$ parameter from Eq. (\ref{eq:loss}) controls the spread and defines the width of separation between the actions in the embedding space $\varepsilon$. Figure~\ref{fig:effect_eta} provides visualization of $2^6$ actions into 2D embedding space. We can clearly observe that increasing the value of $\eta$ up to a certain point (0.01), increases the separation between the points while retaining the structure of those embeddings. Beyond $0.01$ although there is a higher degree of separation we have lost the inherent structure and relations between different action points. In addition, we have observed that as the number of actions increases the value of $\eta$ for which we get maximum separation exponentially decreases as can be seen in  Figure~\ref{fig:eta_agg}.

 \section{Results: Recommender Systems}

We present results from a recommender system task as a second experiment. The aim is to build a recommendation engine which has the ability to suggest meaningful products that result in actual purchase for the user. The goal is to actively engage the customers resulting in profitable business by suggesting relevant `next purchase' products, which might result in higher revenues for the retailer. At the same time, the recommendations need to be personalised to the activity and characteristics of particular users in order to be effective. We capture the essence of this task by asking the RL agent to predict the next product that the user purchased in a real-world data set, and computing the effectiveness by measuring the similarity of embeddings.

\subsection{Data description}

A real world data set was obtained from a large global fashion brand\footnote{Other details redacted for anonymity.}, consisting of two years of product purchases in the apparel category. The raw data includes customer, product and transaction details. The customer information includes age, marital status, and a unique ID, while the product information includes product ID, colour, brand, category, and other details. Additionally, event data was used which includes transaction timestamps, purchased product prices, customer ID, product ID, and product ratings. We curated the data set as per the following description.

In the first step, we discarded all transactions for customer who had fewer than 20 transactions over the course of 2 years. Similarly, we also discarded transactions for products with fewer than 25 transactions over 2 years. The resulting dataset had 10,528 distinct customers and 3,695 unique products. We then prepared an offline dataset divided into states and actions, where the states include customer details (scalar inputs for values such as age, one-hot inputs for categorical variables such as gender or marital status) and features of two consecutive product purchases (continuous or one-hot encoding as applicable). The resulting state vector is of size 242. The `action' consisted of predicting the next (third) product that the customer would purchase. In cases where multiple products were purchased in a single basket, we randomly shuffled their order. Thus curated, we trained all algorithms on a dataset consisting of 1,98,800 transactions. 

\subsection{Baselines and RL training}

The states and actions are available in the offline dataset as described above. The training of the encoder-decoder model is thus straightforward and was described in Section \ref{sec:method}. Note that the action embeddings in this scenario are the projections of product features onto a 2-dimensional real number space. 

The RL agent used in our experiments is DDPG as before, which gives the output in a 2-dimensional embedding space. The reward is determined by the cosine similarity between the predicted action (DDPG output) and the actual purchased item embedding vector obtained from the encoder. It is assumed that giving the recent purchase history of the customer provides context for the model to predict the right product to recommend. Each episode consists of 200 sample transactions randomly sampled from the created offline dataset. The DQN agent as a baseline is not included since it produces values for a set of discrete actions, whereas in this scenario we are computing rewards directly in 2D continuous space. Hence, we consider PG-RA~\cite{chandak2019learning} and JSAE~\cite{joint_training} as baselines. 

\begin{figure}[h]
	\centering
	\includegraphics[width=0.5\textwidth, height=0.25\textheight]{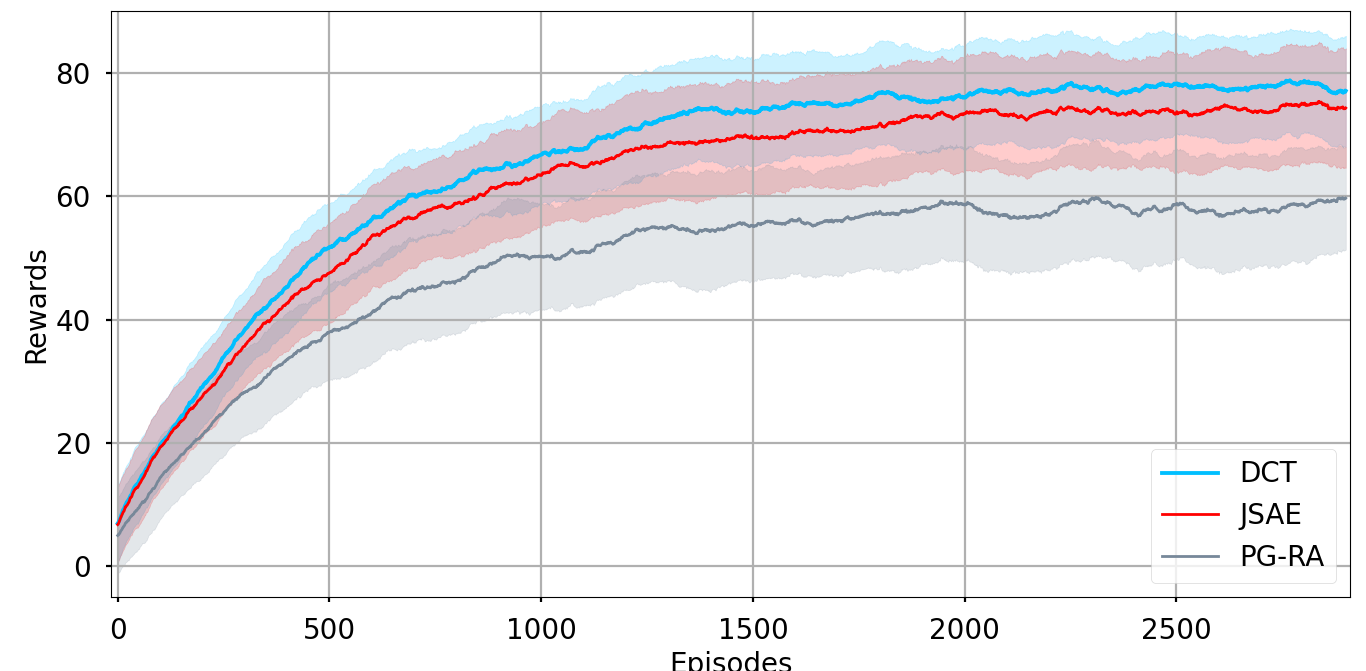}
	
    \caption{Training curves for the proposed method (DCT) and two baselines, over 5 random seeds. DQN is not included since rewards are being computed in 2D continuous space, based on cosine similarity between produced embeddings and ideal action embeddings.}
    \label{fig:rec_agg}
\end{figure}

%\begin{table}[h]
%\caption{Training Results Summary: We observe the mean and standard deviation of final reward for three subsets of the training data set. While DCT is slightly worse JSAE for products with high number of samples, it significantly outperfoms both baselines in subsets with low to medium number of samples. We hypothesize that this behaviour is because of better action embeddings produced by DCT.}
%\label{tab:test_res}
%\resizebox{0.5\textwidth}{!}{%
%\begin{tabular}{|l|ll|ll|ll|}
%\hline
%Transaction    & \multicolumn{2}{c|}{PG-RA}                   & \multicolumn{2}{c|}{JSAE }          & \multicolumn{2}{c|}{DCT}             \\
%           frequency          & \multicolumn{1}{c|}{$\mu$} & \multicolumn{1}{c|}{$\sigma$} & \multicolumn{1}{c|}{$\mu$} & \multicolumn{1}{c|}{$\sigma$} & \multicolumn{1}{c|}{$\mu$} & \multicolumn{1}{c|}{$\sigma$}    \\ \hline

%High            & \multicolumn{1}{l|}{49.54}  & 1.07                   & \multicolumn{1}{l|}{\textbf{72.38}} & 1.37                    & \multicolumn{1}{l|}{\textbf{67.60}}  & 1.24                                \\

%Medium           & \multicolumn{1}{l|}{68.52} & 0.36                   & \multicolumn{1}{l|}{81.645} & 0.38                  & \multicolumn{1}{l|}{\textbf{88.33}} & 1.05                          \\

%Low           & \multicolumn{1}{l|}{36.91}  & 1.62                   & \multicolumn{1}{l|}{56.99} &  0.50                  & \multicolumn{1}{l|}{\textbf{67.06}} & 0.26                                \\ \hline

%\end{tabular}%
%}
%\end{table}

\begin{table}[h]
\centering
\caption{Training Results Summary: We observe the mean and standard deviation of final reward for three subsets of the training data set, split by the number of transactions (frequency of observation). DCT is able to outperform the baselines in all the three cases. We hypothesize that this behaviour is because of better action embeddings produced by DCT.}
\label{tab:test_res}
%\resizebox{0.5\textwidth}{!}{%
\begin{tabular}{|l|ll|ll|ll|}
\hline
Transaction    & \multicolumn{2}{c|}{PG-RA}                   & \multicolumn{2}{c|}{JSAE }          & \multicolumn{2}{c|}{DCT}             \\
           frequency          & \multicolumn{1}{c|}{$\mu$} & \multicolumn{1}{c|}{$\sigma$} & \multicolumn{1}{c|}{$\mu$} & \multicolumn{1}{c|}{$\sigma$} & \multicolumn{1}{c|}{$\mu$} & \multicolumn{1}{c|}{$\sigma$}    \\ \hline

High            & \multicolumn{1}{l|}{60.66}  & 0.12                   & \multicolumn{1}{l|}{84.29} & 3.18                    & \multicolumn{1}{l|}{\textbf{86.63}}  & 1.52                                \\

Medium           & \multicolumn{1}{l|}{59.09} & 0.19                   & \multicolumn{1}{l|}{74.35} & 0.51                  & \multicolumn{1}{l|}{\textbf{78.53}} & 0.65                          \\

Low           & \multicolumn{1}{l|}{56.46}  & 0.37                   & \multicolumn{1}{l|}{73.55} & 0.16                 & \multicolumn{1}{l|}{\textbf{78.31}} & 0.26                                \\ \hline

\end{tabular}%
%}
\end{table}

\subsection{Training Results}

We can observe from Figure \ref{fig:rec_agg} that DCT outperforms other baselines. Since the curves for DCT and JSAE are close to each other, we investigate the results in more detail in Table \ref{tab:test_res}. The table splits the training data set into three types of action labels (products which should be suggested as output). These are products with high number of transactions (highest percentiles), medium, and the fewest transactions. As seen from the table, DCT is able to outperform both the baselines PG-RA \cite{chandak2019learning} and JSAE \cite{joint_training} in all the three subsets on average, with at least one standard deviation of difference from JSAE. Note that as the transaction frequency reduces, the advantage of DCT over JSAE increases. We hypothesize that this is because it is able to learn cleaner representations, as opposed to JSAE which is possibly overfitting to common data points.

%We can observe from figure \ref{fig:rec_agg} that DCT outperforms other baselines. Table \ref{tab:test_res} presents the training results in greater detail, split into users with high number of transactions (highest percentiles), medium, and the fewest transactions.  As seen from the table, the DCT reward is slightly lower than JSAE (but higher than PG-RA) for customer transactions that had high frequency in the dataset. However, DCT significantly outperforms both baselines on low and medium frequency transactional data. We hypothesize that this is because it is able to learn cleaner representations, as opposed to JSAE which is possibly overfitting to common data points. 

\begin{figure}[t]
	\centering
	\includegraphics[width=0.45\textwidth]{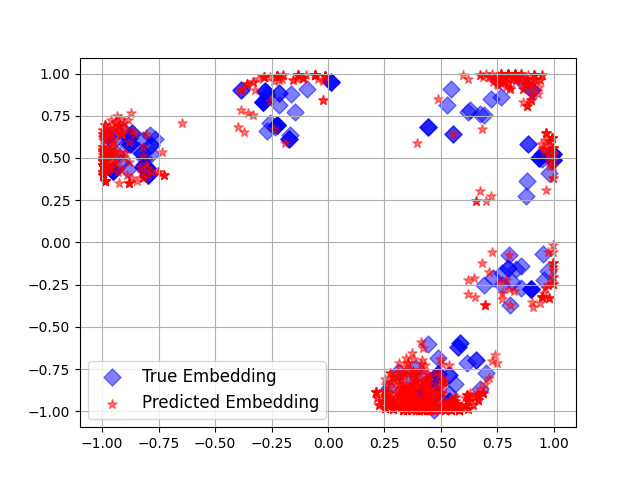}
    \caption{Embedding vectors of products of hitorical transactions of a single customer with 200 transactions. The blue diamonds represent the embeddings of the actual products purchased, while the red stars represent the embeddings of the products with the top 5 softmax scores after decoding the DDPG output.}
	\label{fig:rec_agg1}
\end{figure}

Additional intuition about DCT output is provided by Figure \ref{fig:rec_agg1}, which plots the history of a single customer with 200 transactions in the dataset. The blue diamonds represent the embeddings of the actual 200 products purchased by the customer. The red stars are the embeddings of the 5 products per sample, with the highest softmax scores after decoding the DDPG output. We observe that the placement in the embedding space, as well as the frequencies in each cluster, are approximately matching.

Finally, the fact that we train a specific action decoder that directly outputs softmax scores over all actions, is a great advantage in the deployment phase of such an algorithm. Both baselines \cite{chandak2019learning}, \cite{joint_training} require a k-nearest-neighbour search to compute the actual action, which can be inefficient for a real-time scenario such as a recommendation system. 
%This is due to the decoder function which is trained simultaneously which resulted in better action embedding space. Since, a decoder is already trained, the top 5 ranked products for a customer can be easily obtained from the output of decoder( which directly takes the output of DDPG and provides a probability vector for all the products). This is faster in compared to the baselines where these algorithms have to perform the nearest neigbhor search to provide top ranking recommendations. It is also important to note that as number product increases, time taken for knn search also increases in logarthmic scale.

\section{Conclusion}

Sample efficiency and handling large action spaces are two of the major challenges that are hindering the adoption of RL algorithms into real-world application. PG-RA~\cite{chandak2019learning} and JSAE~\cite{joint_training} have shown how to condense the large action spaces into dense low dimensional embeddings. DCT is an extension of JSAE where along with the prediction of $s_{t+1}$ it also learns to reconstruct actions resulting in richer and robust embeddings. This in turn helps in learning the policy over embedding space in a sample efficient manner.

We can conclude from the experiments that DCT is able to learn across a different number of actions consistently. This is validated across 2 diverse environment of navigation and recommender systems. As a part of the investigation, we have also looked at how loss coefficient $\eta$ affect the structure of embedding.

\bibliographystyle{ACM-Reference-Format} 
\bibliography{references}

\end{document}